\documentclass[letterpaper, 10 pt, conference]{ieeeconf}   
\IEEEoverridecommandlockouts
\usepackage{amsthm}
\usepackage{stmaryrd}
\usepackage{amsmath}
\usepackage{amssymb}
\usepackage{siunitx}

\usepackage{graphicx}
\usepackage{caption}
\usepackage{subcaption}
\usepackage[table]{xcolor}
\usepackage{color}
\usepackage{svg}

\usepackage{multirow}
\usepackage{wrapfig,lipsum,booktabs}
\usepackage{makecell} 

\usepackage{lipsum} 
\usepackage{pdfpages}
\usepackage{kotex}
\usepackage{xurl}

\DeclareMathOperator*{\argmax}{argmax}

\title{\LARGE \textbf{
SGGNet$^2$: Speech-Scene Graph Grounding Network \\ for Speech-guided Navigation 
}}

\author{Dohyun Kim*, Yeseung Kim*, Jaehwi Jang*, Minjae Song*, Woojin Choi, and Daehyung Park\textsuperscript{\textdagger}
    \thanks{*These authors contributed equally to this work. }
    \thanks{All authors are with the Robust Intelligence and Robotics (RIRO) laboratory, Korea Advanced Institute of Science and Technology, Korea ({\tt\small \{dohyun141, yeseung.kim.88, wognl0402, smj0398, cwwojin, daehyung\}@kaist.ac.kr}). {\textsuperscript{\textdagger}}D. Park is the corresponding author. This work was supported by the National Research Foundation of Korea (NRF) grants funded by the Korea government (MSIT) (No. 2021R1A4A303283413 and 2021R1C1C100436813) and the KAIST Convergence Research Institute Operation Program.}
}

\begin{document}

\maketitle
\thispagestyle{empty}
\pagestyle{empty}

\begin{abstract}
The spoken language serves as an accessible and efficient interface, enabling non-experts and disabled users to interact with complex assistant robots. However, accurately grounding language utterances gives a significant challenge due to the acoustic variability in speakers' voices and environmental noise. In this work, we propose a novel speech-scene graph grounding network (SGGNet$^2$) that robustly grounds spoken utterances by leveraging the acoustic similarity between correctly recognized and misrecognized words obtained from automatic speech recognition (ASR) systems. 
To incorporate the acoustic similarity, we extend our previous grounding model, the scene-graph-based grounding network (SGGNet), with the ASR model from NVIDIA NeMo. We accomplish this by feeding the latent vector of speech pronunciations into the BERT-based grounding network within SGGNet. We evaluate the effectiveness of using latent vectors of speech commands in grounding through qualitative and quantitative studies. We also demonstrate the capability of SGGNet$^2$ in a speech-based navigation task using a real quadruped robot, RBQ-3, from Rainbow Robotics. 

\end{abstract}
\section{Introduction}

As the population ages, the demand for caregivers to assist with direct and indirect daily living activities is increasing. 
A potential solution to this challenge is the deployment of robotic caregivers, which can provide increased independence for the elderly and complement human caregivers~\cite{park2018multimodal,kapusta2015task,kapusta2019system}. However, the command interfaces of robots typically require professional knowledge to operate effectively. To improve the usability of robotic assistants, we need a more intuitive interface for non-expert users. 

Natural language serves as one of the most convenient communication interfaces for humans. In robotics, natural language grounding (NLG) refers to the ability of a robot to understand linguistic instructions by establishing a connection between phrases or words in human language and the physical world. Researchers have introduced a wide variety of human instruction grounding approaches~\cite{nyga2018grounding, howard2022intelligence, kim2023natural, roy2019leveraging, arkin2020real}. Classic methods involve identifying the structure and meaning of linguistic instructions~\cite{tellex2011understanding}. Recent approaches utilize neural language models to predict the target actions or objects in the robots' world model~\cite{brohan2023can}. However, the grounding performance inherently depends on the accuracy of converting human speech into text. Any errors in this conversion, caused by speaker variability or environmental noise, significantly degrade the grounding performance, thereby affecting the usability and safety of the robots.

\begin{figure}[t]
    \centering
    \includegraphics[width=\columnwidth]{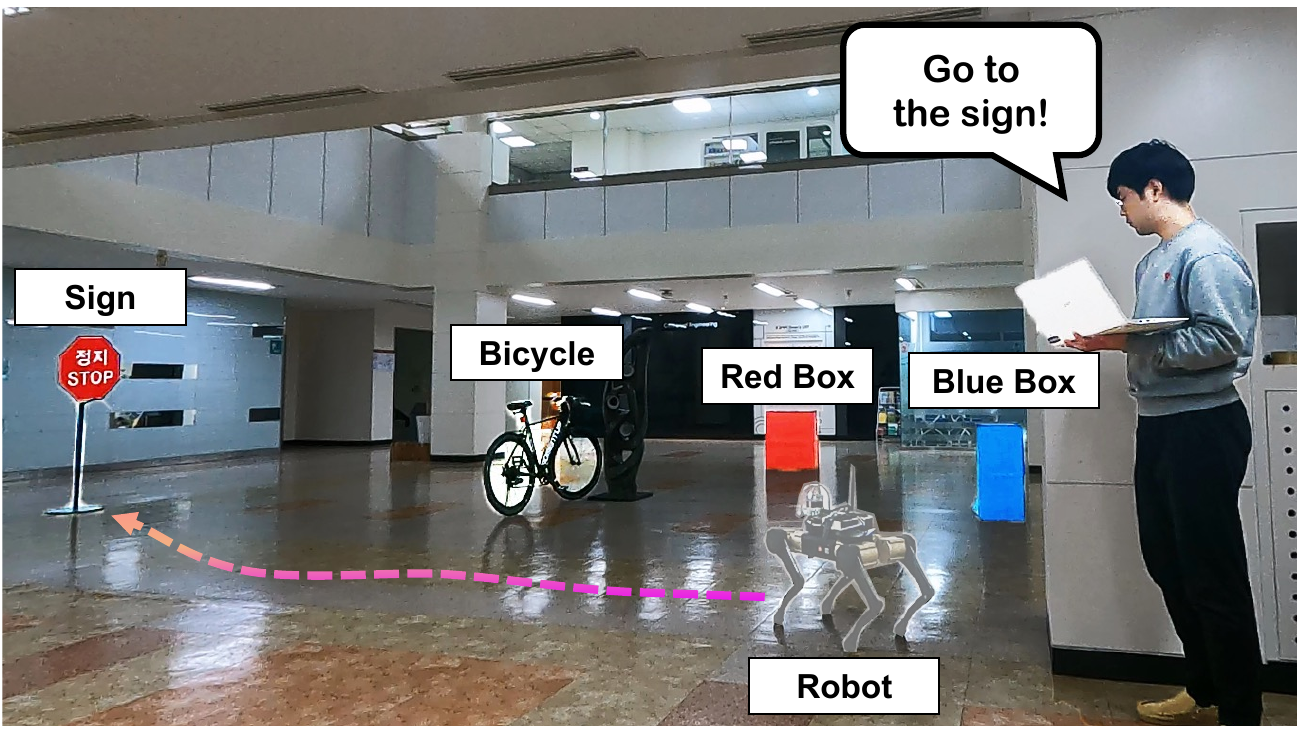} 
    \caption{A capture of speech-based grounding-and-navigation experiment. Our speech-scene graph grounding network (SGGNet$^2$) identifies a target object by associating the latent information of a voice command with a constructed scene graph.}
    \vspace{-1.2em}
    \label{fig:main}
\end{figure}

Automatic speech recognition (ASR), also known as speech-to-text (STT), is the process of converting spoken language into written text. In robotic applications, the STT service is to convert an operator's utterance into a written form, enabling the connection between human intention and the corresponding robot action. Traditional approaches often construct separate generative language, pronunciation, and acoustic models to form a complete ASR pipeline. Recent advancements in  neural network-based methods, such as LAS~\cite{chan2016listen} and NVIDIA-NeMo ASR~\cite{nvidianemo}, have facilitated the joint learning of all ASR components in an end-to-end fashion~\cite{shillingford19_interspeech}. 
The end-to-end schemes offer advantages in training simplicity and compatibility when integrated with other networks to achieve specific task objectives. In this work, we leverage the capabilities of state-of-the-art ASR models to enhance the performance of speech-guided navigation tasks in robotics.

We introduce an end-to-end scheme for the speech-scene graph grounding network (SGGNet$^2$), which robustly identifies target objects in a robot's world model given voice commands. The core idea of SGGNet$^2$ is that words with similar acoustic properties can be generated from closely related latent encodings. By transferring the encoding to the scene-graph grounding network (i.e., SGGNet~\cite{kim2023natural}), we enable the robot correctly understand the intended word that can be vocally misconverted by the ASR system.

We evaluate the proposed grounding network by training it with a Korean navigation dataset. We particularly show the similarity between the acoustic properties of words in the latent space. This unified network significantly improves grounding performance compared to conventional frameworks. Furthermore, we demonstrate the speech-guided navigation with SGGNet$^2$ using a quadruped robot, RBQ-3, toward assistive navigation or fetching tasks (see Fig.~\ref{fig:main}). Our work ultimately enhances the capability of robotic assistance through natural language commands, benefiting non-expert users.

Our contributions are as follows:
\begin{itemize}    
    \item we propose an end-to-end structure for the speech-scene graph grounding network, which resolves the issue of misconversion from ASR, 
    \item we introduce a Korean speech-based grounding framework for training navigation tasks, utilizing NVIDIA-NeMo ASR with the Korean navigation dataset, and
    \item we perform a real-world demonstration by deploying SGGNet$^2$ on a real quadruped robot.
\end{itemize}

\section{Related Work}

\subsection{Natural-language grounding}

Natural language grounding has become increasingly prominent in the field of robotics, allowing robots to follow verbal instructions without the need for a specialized interface. Early studies primarily focused on identifying target objects within the physical environment~\cite{achlioptas2020referit3d} or specifying target locations by leveraging on spatial relationships present~\cite{chen2020scanrefer} in written instructions. More advanced works began considering historical referring expressions~\cite{roy2019leveraging}, background knowledge~\cite{nyga2018grounding,chen2020enabling}, and the affordances of various skills~\cite{brohan2023can}. Other approaches utilize graphical models to represent natural language utterances through syntactic parse trees of phrases~\cite{tellex2011understanding, howard2014natural}. However, syntactic parsing is often vulnerable to the ambiguity of unstructured expressions, which can undermine grounding performance.

Recent developments have employed language models such as OpenAI's GPT~\cite{radford2018improving} and BERT~\cite{kenton2019bert}, which have substantially improved grounding performance. These models possess rich commonsense intent representations capable of handling the variability of linguistic commands. Researchers have also explored aligning natural language with geometric trajectories~\cite{bucker2022reshaping} or integrating human feedback~\cite{thomason2016learning} to enhance grounding performance. However, these works still rely on the input text converted from the user's speech. In contrast, our proposed framework merges speech and perceptual data to improve robustness against acoustic variability.

\subsection{Automatic speech recognition for language grounding}

To deliver our intentions to robots, researchers have used off-the-shelf ASR. Traditional ASR systems often use isolated word entries with classic machine learning models such as hidden Markov models (HMMs). In order to accommodate a wide variety of continuous speech, commercial entities offer cloud-based ASR systems: Microsoft Speech Platform SDK~\cite{ms_STT} and Google Cloud Speech API~\cite{google_stt}. 
While these software systems provide high accuracy, various factors like the user's accent, linguistic variation, and background noise can significantly affect the precision of transcription.

The advance of neural network models has improved the performance of standalone ASR systems.
For example, NVIDIA NeMo~\cite{nvidianemo}, a representative conversational AI toolkit, enables a robot to output text from an acoustic model for speech recognition. 
In this work, we not only utilize the capability of the ASR model but also extract speech features to improve grounding performance in noisy real-world settings.

\section{Problem Formulation}

\begin{figure*}[t]
    \centering
    \includegraphics[width=0.95\textwidth]{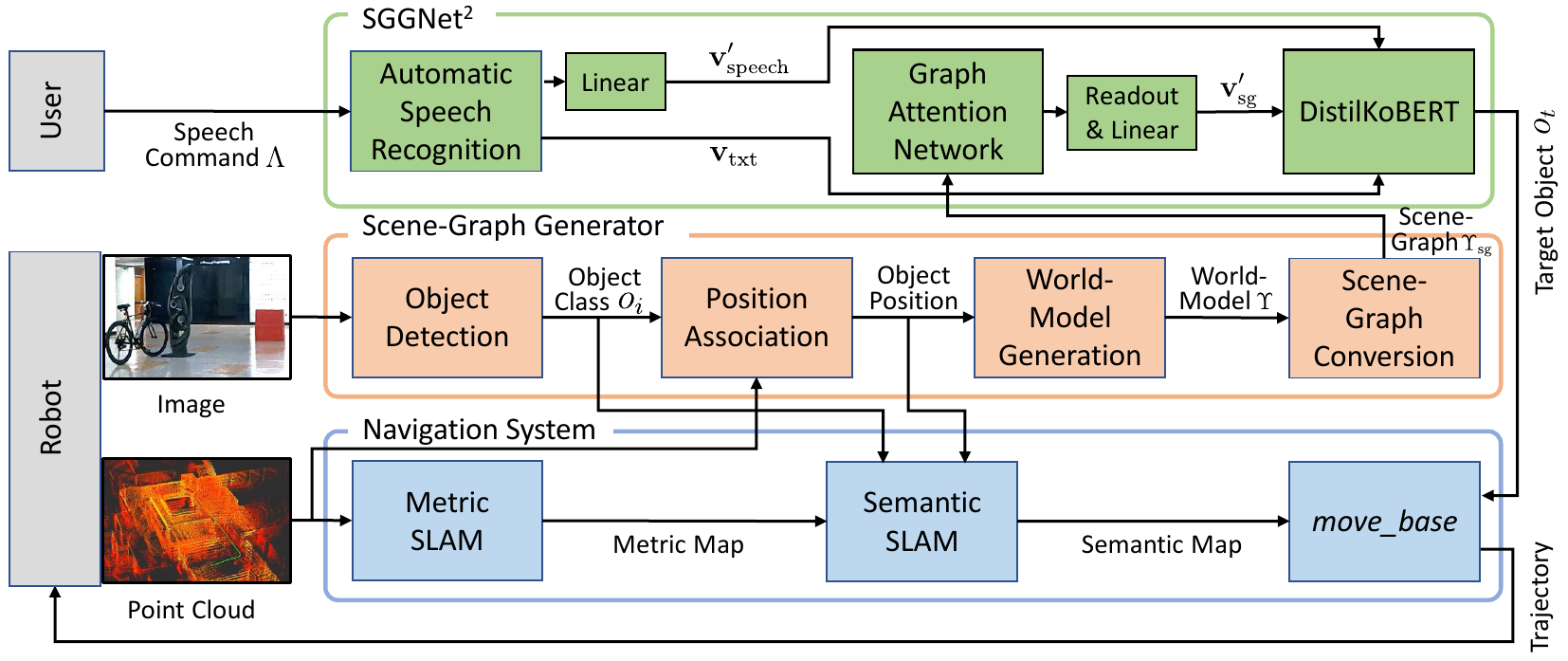} 
    \caption{The overall architecture of a speech-enabled navigation system incorporating SGGNet$^2$. The system receives a user's speech command along with perceptual information from the robot, including RGB images and point cloud data. Through SGGNet$^2$, the robot identifies a target object (i.e., destination). Subsequently, our navigation system executes the user's command, guiding the robot toward the specified destination.}
    \vspace{-1.5em}
    \label{fig:overview}
\end{figure*}

Consider a mobile robot operating in a workspace with a set of detectable objects $o_i \in \mathcal{O}$. Our goal is to enable the robot to ground a target object $o_t\in \mathcal{O}$ that satisfies a human's speech command $\Lambda$, given a world model $\Upsilon$. The robot receives the spoken command $\Lambda$ through a natural language interface (i.e., microphone).  The world model $\Upsilon$ includes information such as the class, position, and other recognizable features of each observable object $o_i \in \mathcal{O}$. The robot recognizes each object using a detector $\varrho$. Conventional language-grounding framework finds the target object $o_t$ using modularized ASR, world model, and natural language grounding functions, assuming the conditional independence of visual and auditory information:
\begin{align}
    o_t^* &= \argmax\limits_{o_t\in\mathcal{O}} P(o_t | \Lambda, \mathcal{O}) \nonumber\\
    &= \argmax\limits_{o_t\in\mathcal{O}} P(o_t | \mathbf{Z}, \Lambda, \Upsilon, \mathcal{O}) P(\mathbf{Z} | \Lambda, \Upsilon, \mathcal{O}) P(\Upsilon| \Lambda, \mathcal{O}) \nonumber\\    
   &= \argmax\limits_{o_t\in\mathcal{O}} \underbrace{P(o_t | \mathbf{Z}, \Upsilon)}_{\substack{\text{Natural language} \\ \text{grounding}}} \underbrace{P(\mathbf{Z} | \Lambda)}_{\substack{\text{Automatic}\\\text{speech recognition}}} \underbrace{P(\Upsilon| \mathcal{O})}_{\substack{\text{World-model}\\\text{generator}}},
   \label{eq_past_nlg}
\end{align}
where $\mathbf{Z}$ is the written utterance that corresponds to $\Lambda$. However, the acoustic variability and noise of spoken utterances in the real world often degrade the performance of ASR.

We make the assumption that words or characters that sound similar result in vectors that are close to each other in the latent space. This can help improve grounding performance in cases where the text is misconverted. For example, the Korean word ``\textit{자전거} ([\textit{jajeongeo}], meaning \textit{bicycle})" is often misconverted to ``\textit{하얀 거} ([\textit{hayangeo}], meaning \textit{white one})" or ``\textit{다른 거} ([\textit{dareungeo}], meaning \textit{something else})" when input to the ASR model. This can degrade the grounding process, especially in a noisy environment.

We improve the framework with Eq.~(\ref{eq_past_nlg}) by formulating an end-to-end scheme of grounding network, SGGNet$^2$, directly grounding $o_t$ given the spoken utterance and a scene-graph  $\Upsilon_{\text{sg}}$, 
\begin{align}
    o_t^* &= \argmax\limits_{o_t\in\mathcal{O}} \underbrace{P(o_t | \Lambda, \Upsilon_{\text{sg}})}_{\text{SGGNet$^2$}} \underbrace{P(\Upsilon_{\text{sg}}| \mathcal{O})}_{\substack{\text{Scene-graph}\\\text{generator}}},
\end{align}
where the scene-graph $\Upsilon_{\text{sg}}$ is a graph-based representation of the world model encoding the semantic relationships between entities based on our past work~\cite{kim2023natural}. For computational convenience, our model first calculates the best scene-graph $\Upsilon^*_{\text{sg}}$ and then computes the likelihood of the target object given $\Lambda$ and $\Upsilon^*_{\text{sg}}$.  
\section{Speech-Scene Graph Grounding Network}

\subsection{Overview of SGGNet$^2$ }
The speech-scene graph grounding network, SGGNet$^2$, is a unified structure of the speech-grounding model that combines a scene-graph grounding network, SGGNet~\cite{kim2023natural}, with an ASR network from NVIDIA NeMo ASR~\cite{nvidianemo}.
Unlike SGGNet, which accepts text as input, SGGNet$^2$ directly processes human-speech commands. Fig.~\ref{fig:overview} shows the overall architecture of the speech-enabled navigation system with SGGNet$^2$, which grounds a target object in a scene-graph obtained from an RGB-D camera given a human-speech command. 

SGGNet$^2$ particularly uses a Conformer-CTC \cite{conformer} model via NeMo ASR, which converts audio input $\Lambda$ to textual output $\mathbf{Z}$. To use a variety of languages, we pre-train the model with the Korean speech dataset (see Sec.~\ref{ssec:korean_asr}) before training the entire SGGNet$^2$ model with the navigation command dataset. We input the embedded speech vector $\mathbf{v}_{\text{speech}}\in \mathbb{R}^n$, the output text vector $\mathbf{v}_{\text{txt}}\in \mathbb{R}^m$, and an embedded vector of the scene-graph $\mathbf{v}_{\text{sg}}\in \mathbb{R}^k$ (see Sec.~\ref{ssec:sg}) to DistilKoBERT~\cite{park2019distilkobert}, for grounding a target object, where $n$, $m$, and $k$ are user-defined sizes of input vectors.

\subsection{Automatic Korean speech recognition} \label{ssec:korean_asr}
We develop an automatic Korean-speech recognition network using NVIDIA NeMO \cite{nvidianemo}, which is a conversational AI toolkit with prebuilt modules for ASR, natural language processing (NLP), and TTS synthesis. We use NeMo's Conformer-CTC-based ASR model, a variant of the Conformer, which has an encoder using CTC loss~\cite{graves2006connectionist} and a linear decoder. This structure gives a non-autoregressive model enabling fast inference.

To enable the ASR model to recognize Korean speech, we pre-train it with the KsponSpeech dataset, a large-scale corpus of spontaneous Korean speech~\cite{app10196936}. This dataset includes $620,000$ utterances of PCM format recordings ($965.2$ hours) and text-Korean phonetic transcriptions from $2,000$ speakers, with $54\%$ women and $46\%$ men. We preprocess the dataset following the guidelines outlined in \cite{2021-kospeech}, converting the PCM recordings to WAV and limiting the vocabulary to $5,000$ words. 
We apply a \SI{30}{\decibel} threshold to remove silence sections in the speech. For transcriptions, we eliminate special tokens indicating background noise, breathing, and other unwanted elements. Additionally, we process the transcriptions into characters, ensuring a suitable format for the model's further analysis.

We pre-train the model for $20$ epochs with a batch size of $32$ with Adam optimizer setting $10^{-6}$ of weight decay. To increase the dataset size, we employed SpecAugment~\cite{Park_2019} for data augmentation.

\subsection{Scene-graph generator}\label{ssec:sg}
We design a scene-graph generator that constructs a scene-graph $\Upsilon_{\text{sg}}$ given visual information. The generator consists of four steps: object detection, position association, world-model generation, and scene-graph conversion. 
To detect environmental objects around the robot, we use an object detector, YOLO~\cite{redmon2016you} that subscribes to RGB image streams and finds bounding boxes estimating their positions in pixel coordinates.
We use an adaptive clustering algorithm LiDAR point clouds \cite{yan2020online} to associate the detected 2-D objects with 3-D box clusters to obtain their 3-D positions ($\in \mathbb{R}^3$, see details in \cite{kim2023natural}).
We then add any new detection result to an object dictionary (i.e., world model $\Upsilon$) with the object class, attribute, and position.
The system stores the class and attribute of each object in the textual format while representing the position using a three-dimensional vector in $\mathbb{R}^3$.
Then, to represent the spatial relationships between objects, we convert the world model dictionary to a scene-graph, $\Upsilon_{\text{sg}}=(\mathcal{V},\mathcal{E})$, where $\mathcal{V}$ and $\mathcal{E}$ are a set of object nodes and directional edges between them, respectively.
Pre-defined predicates, \textit{left}, \textit{right}, \textit{front}, and \textit{behind}, compose the edge attributes between two interconnected nodes.
A rule-based approach is used to ascertain these predicates, which considers the $x$ and $y$ coordinates of the nodes within the global frame (see details in \cite{kim2023natural}).
The system stores edge predicates, which represent spatial relationships, in textual format.
When multiple predicates are present, the system combines them using a blank space as a separator, for example, \textit{left behind}.

\begin{figure}[t]
    \centering
    \includegraphics[width=\columnwidth]{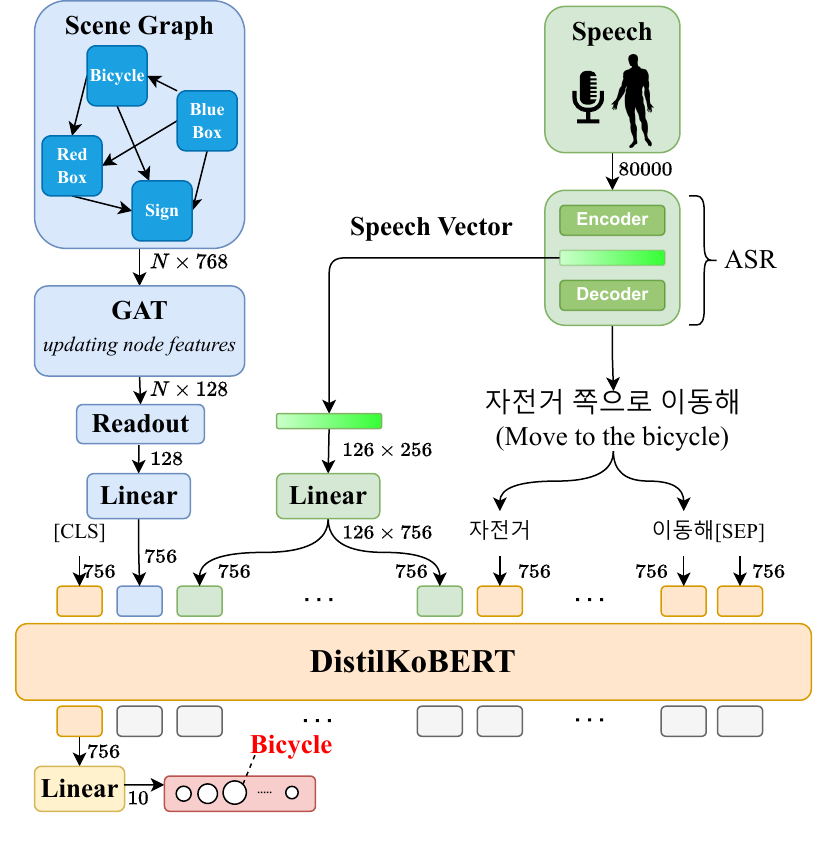}
    \caption{
    The architecture of our proposed speech-scene graph grounding network (SGGNet$^2$). The graph attention network (GAT) generates a scene-graph vector, while the ASR module converts a speech command into text and also extract a latent vector. Finally, the DistilKoBERT processes the scene-graph vector, speech vector, and text to identify the best target object in the world. Each number on the arrow represents the dimension of each vector.
    }
    \vspace{-1.1em}
    \label{fig:SGGNet}
\end{figure}

\subsection{End-to-end grounding network design \& training }
We combine the ASR network as a speech encoder alongside the graph encoder to build an end-to-end structure of SGGNet$^2$. Fig.~\ref{fig:SGGNet} shows the detailed structure, which shows the embedded vectors from the graph encoder and the ASR module, as well as the converted text command are input to the language model, such as BERT, to ground a sentence of speech input to a target entity in the scene-graph. We describe each input below.

\noindent 1) \textbf{Scene-graph vector $\mathbf{v}_{\text{sg}}$}: We extract a latent vector $\mathbf{v}_{\text{sg}}\in \mathbb{R}^{k=128}$ of the scene-graph $\Upsilon_{\text{sg}}$ passing that of nodes and edges through a graph attention network (GAT)~\cite{velivckovic2018graph}. In detail, let $N$ be the number of nodes in a fully-connected scene-graph $\Upsilon_{\text{sg}}$. We convert each node feature (e.g., object class, \textit{bicycle}) to a vector $\mathbf{x}_i \in \mathbb{R}^{128}$ encoding it through DistilKoBERT~\cite{park2019distilkobert}, where $i \in \{1, ..., N \}$. Likewise, we also encode each edge feature, which is the spatial relationship (e.g., \textit{left behind}) between $i$-th and $j$-th nodes, to a vector $e_{ij} \in \mathbb{R}^{128}$, where $j \in \{1, ..., N \}$.

We update node features $\mathbf{X}=[\mathbf{x}_1, ..., \mathbf{x}_N ]$ via a GAT function $f_{\text{GAT}}$, where we compute the attention scores $\alpha_{ij}^{(L)}$ from $i$-th node to $j$-th node at $L$-th layer based on \cite{liang2021graphvqa}. Then, we compute the latent vector $\mathbf{v}_{\text{sg}}$ given the node features and the edge features $\mathbf{E}=[e_{12}, ..., e_{ij}, ... ]$ via a GAT function, $f_{\text{GAT}}$:
\begin{align}
 \mathbf{h} &= f_{\text{GAT}}(\mathbf{X}, \mathbf{E}) \\
 \mathbf{v}_{\text{sg}} &= \phi ( \mathbf{h} )
 \label{eq_gcn}
\end{align}
where $\mathbf{h}$ and $\phi$ are nodes' new embedding ($\in \mathbb{R}^{N \times 128}$) and a max-pooling readout function, respectively. Note that we use a ReLU activation function between layers.

\noindent 2) \textbf{ASR speech vector $\bf{v}_{\text{speech}}$}: The ASR model comprises two primary components: an encoder and a decoder.
The encoder converts the speech command $\Lambda$ to a hidden vector $\mathbf{v}_{\text{speech}, t} \in \mathbb{R}^{256}$ that captures the acoustic features of the spoken words at each step $t \in \{0, ..., T \}$ given the spoken input with length $T$.
In this work, we use $T=126$.

\noindent 3) \textbf{ASR output text vector $\bf{v}_{\text{txt}}$}:
The Conformer-CTC's linear decoder predicts a distribution of characters or words given each hidden vector $\mathbf{v}_{\text{speech}, t}$. By selecting the best one for each step, we obtain a sequence of output texts that can be converted to an embedding vector, $\bf{v}_{\text{txt}} \in \mathbb{Z}^{m=756}$.

We finally input the described embedding vectors, $[\mathbf{v}_{\text{sg}}', \mathbf{v}_{\text{speech}}', \mathbf{v}_{\text{txt}}]$, to the language model, DistilKoBERT~\cite{park2019distilkobert} for grounding a target object in the environment, where $\mathbf{v}_{\text{sg}}' \in \mathbb{R}^{756}$ and $\mathbf{v}_{\text{speech}}' \in \mathbb{R}^{756}$ are the outputs of linear layers. Following the standard BERT~\cite{kenton2019bert} convention, we insert \textit{[CLS]} and \textit{[SEP]} tokens at the beginning and end of the input, respectively. We then add an extra linear layer to the language model, using the \textit{[CLS]} token as input for target object classification.

For training, our model takes a speech command and an associated scene-graph as input. We also provide a target object in the environment, as a ground-truth output. We use a cross-entropy loss function and the AdamW~\cite{loshchilov2019decoupled} optimizer, which has a learning rate of $5 \times 10^{-5}$ and weight decay of $0.01$.
We employ a learning rate scheduler that uses cosine annealing with warm restarts ~\cite{loshchilov2017sgdr} during the training process. We train the model around $200$ epochs.

\begin{figure}[t]
    \centering
    \includegraphics[height=5.75cm]{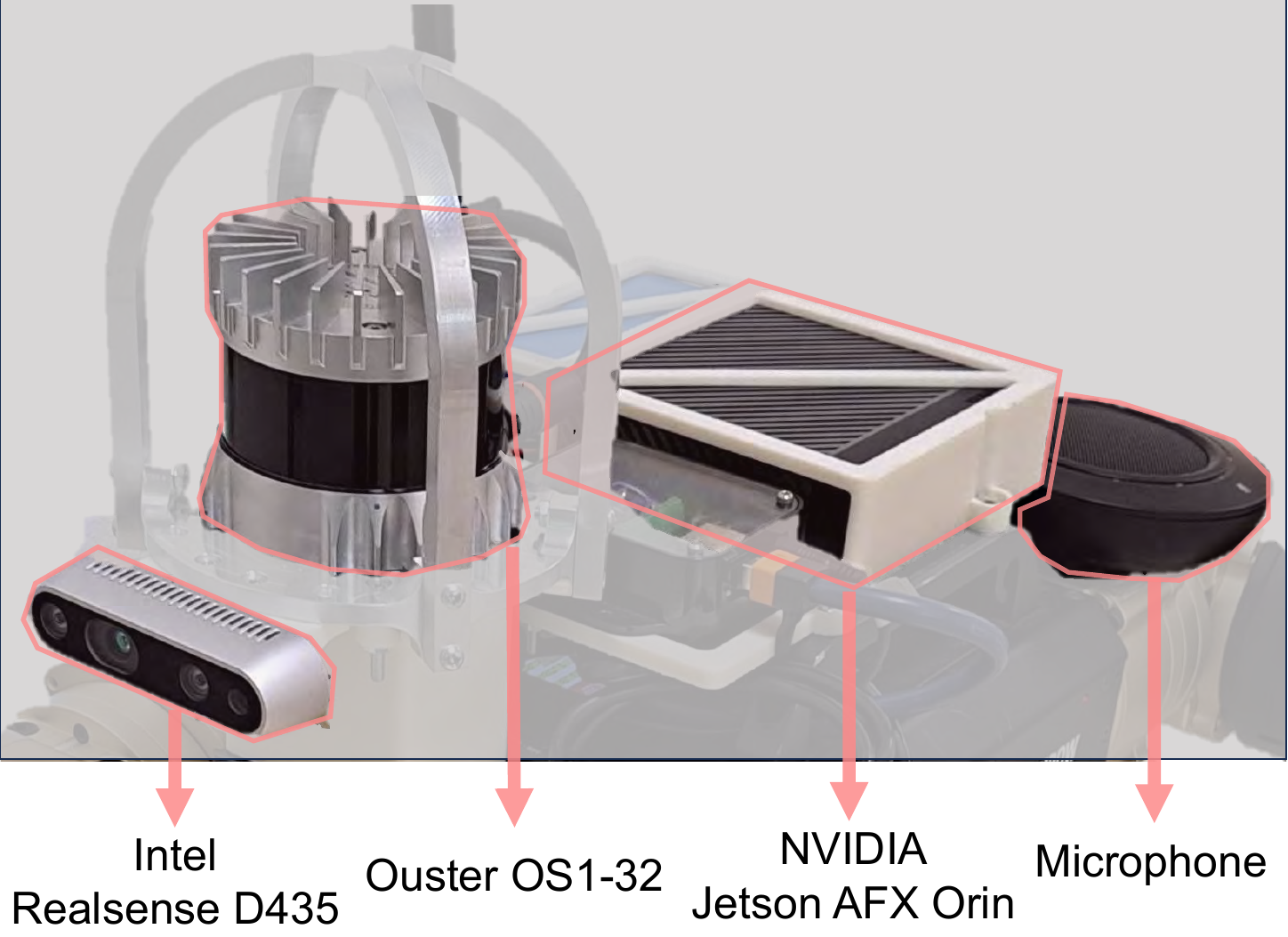} 
    \caption{The hardware setup employed in our study facilitated sensing, computation, metric and semantic map creation, and communication with human operators. To operate the devices within the setup, we utilized the ROS framework.}
    \vspace{-1.1em}
    \label{fig:setup}
\end{figure}

\section{Experimental Setup}

\subsection{Data collection}

We create a speech-scene graph dataset for training SGGNet$^2$ in the robotic navigation domain. In the dataset, each sample contains a human speech, an equivalent text command, and an associated scene-graph. The speeches are a pattern of navigation commands consisting of a \textit{verb},  a \textit{preposition}, and an \textit{object}. For example, we use ``자전거로 가" equivalent to ``Move to the bicycle." To build a dataset, we first selected $10$ classes of target objects: \textit{bicycle}, \textit{sign}, \textit{desk}, \textit{door}, \textit{window}, \textit{red box}, \textit{blue box},  \textit{black box}, \textit{car}, and \textit{suitcase}. We sampled between two and eight synonyms per class. For example, for the object class \textit{table}, we additionally sampled \textit{desk} and \textit{stand}. Finally, we have a total of 35 object names, where each is assigned to one of the training and test sets. Associating to each name, we sampled a list of suitable verbs and prepositions through GPT-4~\cite{GPT-42023OpenAI} given exemplar navigation commands. Finally, we artificially generated $390$ sentences per object by combining a verb, a preposition, and an object name. In other words, we generated a total of $13,650$ text commands that consist of $9,750$ training and $3,900$ test sets. Note that we set the dataset to be independent by assigning non-overlapping object names. 

We also created a scene-graph dataset associated with the text commands. For each command involving a target object, we randomly selected between two and eight additional objects, as nodes, that do not overlap each other. We assigned a 2-dimensional location to each object and fully connect them with edges, representing geometric relationships such as \textit{left}, \textit{right}, \textit{front}, and \textit{behind}. Note that each graph only includes one object per object type. Additionally, each edge can hold multiple relationships, such as `\textit{left behind}.' 

To produce synthetic speech data associated with the commands, we converted the text commands to artificial voice using Naver Clova Voice \cite{kang2018clova}, an advanced text-to-speech (TTS) synthesizer. To maintain diversity in the speech, we randomly selected five male and five female voices. The synthesis system generates a \SI{24}{\kilo\hertz} and 16 bits linear WAV file, which is then padded to be \SI{5}{\second} long.

\begin{figure}[t]
    \centering
    \includegraphics[width=0.95\columnwidth]{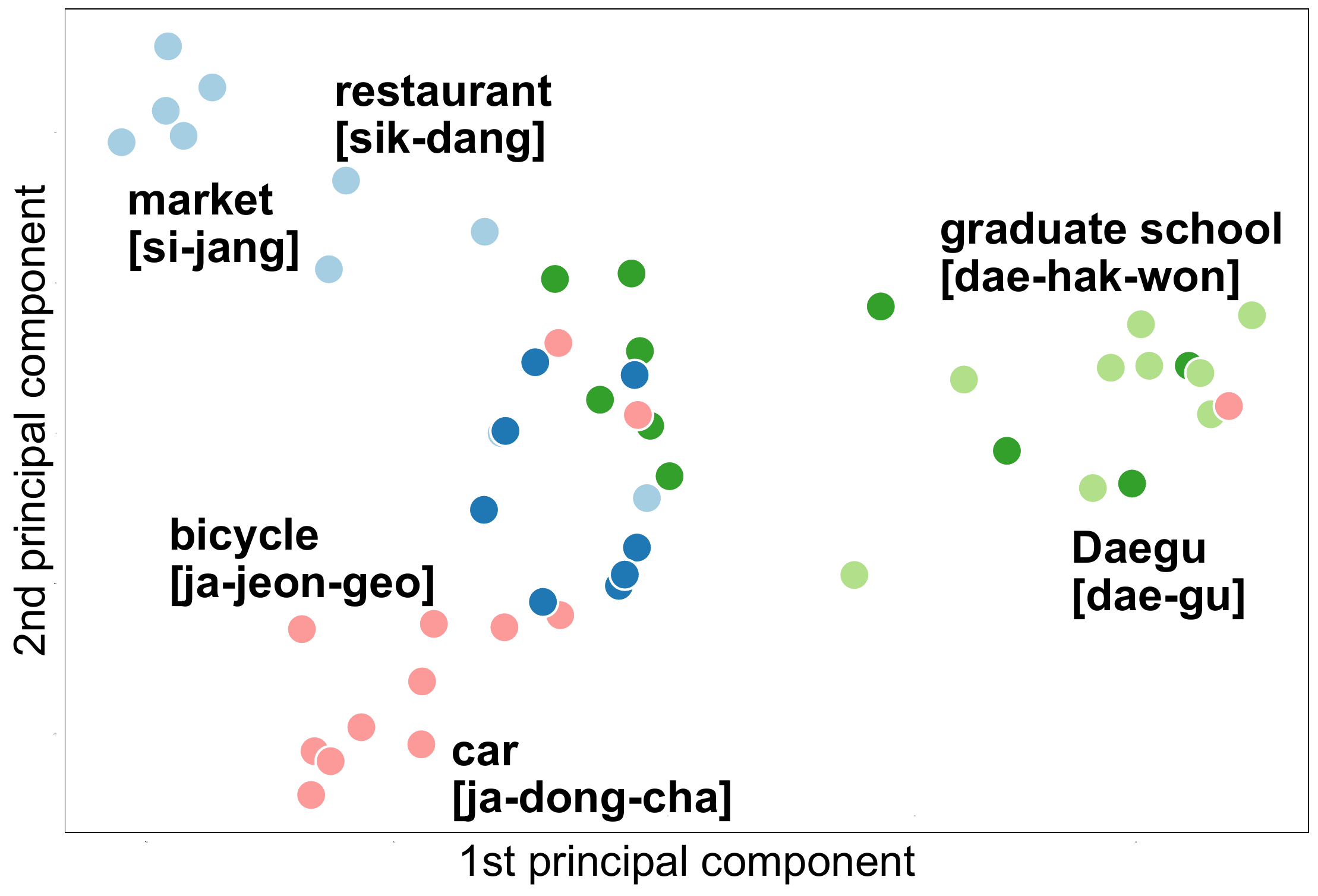} 
    \caption{
    Visualization of the word latent vectors, where we sampled 10 words in each category out of 5 categories (i.e., a total of $50$). The same category of words starts with a particular Hangul consonant.  
    For example, red dots are words starting with `ㅈ.' Blue, sky blue, green, and light green dots represent words starting with `ㅎ,' `ㅅ,' `ㅂ,' and `ㄷ' each.
    We project their latent vectors into 2-dimensional space through principle component analysis.   
    }
    \vspace{-1.1em}
    \label{fig:training}
\end{figure}

\begin{figure*}[t]
    \centering
    \includegraphics[width=\textwidth]{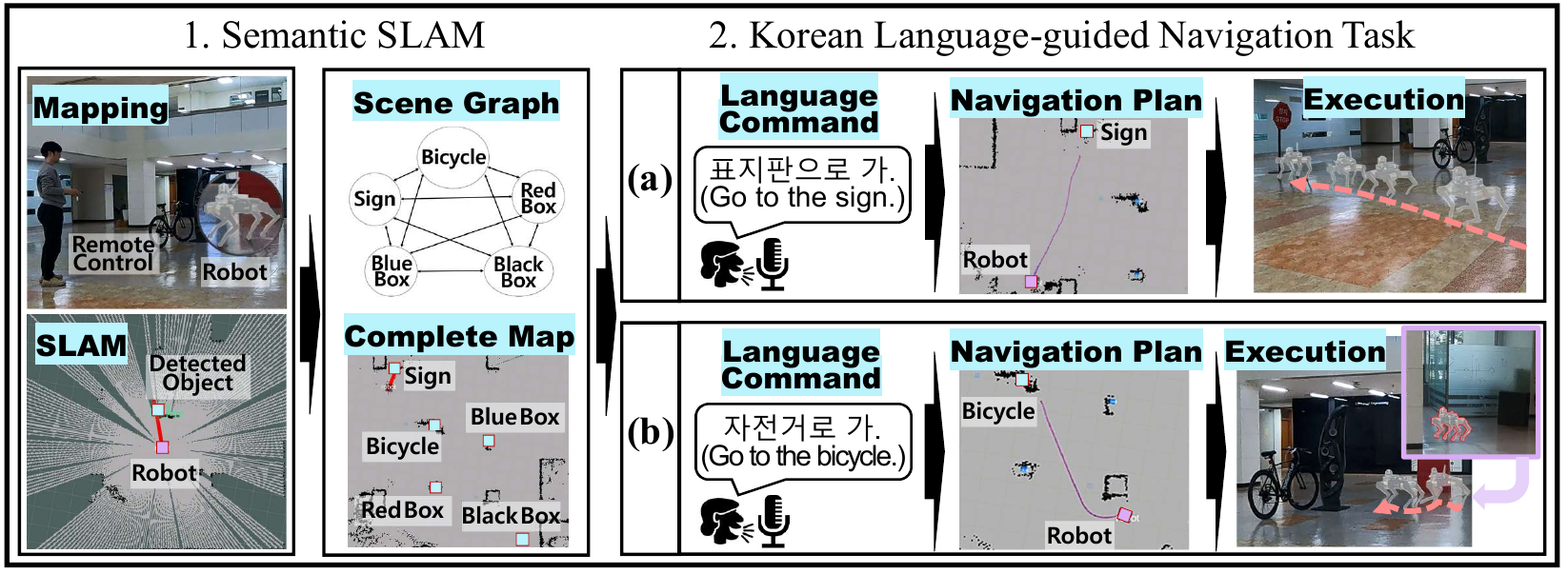} 
    \caption{Demonstrations of the quadruped robot RBQ-3 using the proposed speech-enabled navigation system through the following scenarios: (1) Semantic SLAM, where a human operator controls the robot to generate a map and scene graph, and (2) Korean language-guided navigation following spoken commands. Scenarios include (a) navigation towards a sign and (b) navigation to a bicycle despite occlusion caused by a pillar.}
    \vspace{-1.1em}
    \label{fig:demo}
\end{figure*}

\subsection{Robot setup for speech-enabled navigation}
We briefly introduce our hardware setup and navigation system for semantic navigation. Our robot system consists of a quadruped robot RBQ-3 from Rainbow Robotics, a GPU-enabled single-board computer NVIDIA Jetson AGX Orin, and additional sensors. The RBQ-3 robot is a 12-degree-of-freedom quadruped robot with a maximum payload \SI{3}{\kg}. To perform LiDAR-inertial navigation and semantic map generation, we mount an Ouster OS-1 32-channel LiDAR sensor and a Realsense D435 RGB-D camera, as shown in Fig.~\ref{fig:setup}. In addition, we also mounted a microphone-speaker for speech-based navigation, enabling the robot to communicate with operators. The Jetson computer collects sensor information and controls the robot by sending commands for \textit{speed} and \textit{direction} to the robot's controller. We integrated the software and communication systems using the robot operating system (ROS) with the Noetic version. 

Our navigation system consists of two layers: metric and semantic simultaneous-localization-and-mappings (SLAMs). For the metric SLAM, we construct a dense 3-D point cloud map using a LiDAR-inertial odometry framework, FAST-LIO2 \cite{xu2022fast}, and generate local \& global collision-free trajectories within the map using a \textit{move\_base} package in ROS. Then, for semantic SLAM, our system uses YOLO ~\cite{redmon2016you} and an adaptive point-cloud clustering algorithm \cite{yan2020online} depicted in Sec. \ref{ssec:sg}.
By labeling each cluster, we can create a semantic map which is also used for scene-graph generation and natural language grounding. In other words, when a user can indicate a target object in the world, the robot can recognize its physical location. Finally, we control the robot by sending \textit{speed} and \textit{direction} computed from the semantic navigation system. 

For speech-scene graph grounding, our system receives the voice signal via the microphone-speaker and maintains a scene-graph converted from the detection and localization information of the objects in the semantic SLAM. Finally, when our SGGNet$^2$ returns a target name given the two pieces of information, our navigation system enables the robot to reach the real-world target.

\section{Evaluation}

We hypothesize that the similarity in acoustic properties between misconverted and correct words assists in identifying the correct target object by leveraging their close embeddings in the latent space of ASR. We show the effectiveness of this hypothesis through statistical evaluations and a real-robot demonstration.

\subsection{Statistical evaluation}
Fig.~\ref{fig:training} shows the projection of various latent vectors in 2-D space with their meaning and pronunciation. We found similar pronunciations of words are closely located in the 2-D space regardless of their meanings. For example, although `restaurant [sik-dang]' and `market [si-jang]' are different meanings of words, their latent vectors are close to each other. Thus, regardless of misconversion in ASR, using latent vectors can mitigate the effect of any misconverted word. To obtain the projection points in Fig.~\ref{fig:training}, we sampled $50$ words from ASR's vocabulary list and performed principle component analysis (PCA) of their latent vectors. We particularly visualized the first two components of the PCA results.

Table~\ref{table_eval} shows the ablation study that compares the grounding accuracy of SGGNet$^2$ with variants. Our proposed grounding network SGGNet$^2$ outperformed all the other frameworks with the highest grounding accuracy over $85$\%, which is about $8$\%p higher than SGGNet$^2$ removing the latent speech vector $\mathbf{v}_{\text{speech}}$. The accuracy of SGGNet$^2$ is also about $5$\%p higher than that of the next best method SGGNet with the ground-truth written utterance $\mathbf{Z}^*$ and the same size as $\mathbf{v}_{\text{sg}}$. The results indicate the use of the latent speech vector $\mathbf{v}_{\text{speech}}$ not only resolves the problem of acoustic variability/noise but also improves the grounding performance given the ground truth texts. In addition, noticeable results are that the sole use of ASR-converted written utterances degrades the grounding performance on the previous SGGNet. The overall comparison result indicates that our method, SGGNet$^2$, is robust and effective.

\begin{table}
\centering
\caption{Comparison of grounding networks given a test set of spoken utterances. The input types, $\Lambda$, $\mathbf{Z}$, and $\mathbf{Z}^*$, denote the spoken utterance, written utterance from ASR, and ground-truth written utterance, respectively. The method SGGNet$^2$/$\mathbf{v}_{\text{speech}}$ is a variant of SGGNet$^2$ removing $\mathbf{v}_{\text{speech}}$. }
\label{table_eval}
\resizebox{\columnwidth}{!}{
\begin{tabular}{ccc}
\noalign{\smallskip}\noalign{\smallskip}\Xhline{2\arrayrulewidth}
Method & Input & Grounding accuracy (\%) \\
\hline
\multirow{2}{*}{SGGNet~\cite{kim2023natural}} & $\mathbf{Z}$ & 26.7 \\
& $\mathbf{Z}^*$ & $48.0$ \\
\hline
\multirow{2}{*}{Modified SGGNet} & $\mathbf{Z}$ & $76.0$ \\
& $\mathbf{Z}^*$ & $80.7$ \\
\hline
SGGNet$^2$/$\mathbf{v}_{\text{speech}}$ & $\Lambda$ & $77.0$ \\\hline
SGGNet$^2$ (\textit{Ours}) & $\Lambda$ & $\mathbf{85.3}$ \\
\Xhline{2\arrayrulewidth}
\end{tabular}
}
\vspace{-1.0em}
\end{table}

\subsection{Demonstration with a real robot RBQ-3}
We finally demonstrated the effectiveness of our SGGNet$^2$ in a real-world scenario for speech-enabled semantic navigation tasks. Fig.~\ref{fig:demo} shows two examples of the indoor speech-enabled navigation experiment. A human operator speaks his Korean command, such as ``Go to the red box (빨간색 박스로 가)" or ``Move in front of the bicycle (자전거 앞으로 이동해)," that is delivered to the robot's microphone-speaker or a laptop's built-in microphone the remote operator is holding. Our model, SGGNet$^2$, successfully identified the intended destination out of five candidates and our RBQ-3 robot was able to reach the destination. 
Further, even in the case when the pillar occluded the viewpoint of the robot as shown in Fig.~\ref{fig:demo} (b),  our robot could correctly ground the intended object by leveraging the scene-graph. 
This indicates the effectiveness of scene-graph-based grounding as compared to image-based grounding.
Note that we enabled the robot to construct a 2-D metric/semantic map as well as a scene-graph in advance, placing five objects: a \textit{bicycle}, a \textit{sign}, a \textit{red box}, a \textit{blue box}, and a \textit{black box} in advance.

\section{Conclusion}
We proposed a framework for grounding Korean speech-enabled navigation commands, which combines a scene graph-based grounding network with a Korean acoustic-speech recognition network. Our evaluation shows that similar acoustic properties of words lead to similar encodings, reducing grounding failures and enhancing the robustness of SGGNet. Furthermore, we validate the effectiveness of our approach through a real-world navigation task. We anticipate that our study enhances the accessibility of robotic assistance by enabling natural language commands for non-expert users.

\bibliographystyle{ieeetr}

\bibliography{main}

\end{document}